\documentclass[10pt,journal,compsoc]{IEEEtran}
\usepackage{amsmath,amsfonts}
\usepackage{algorithmic}
\usepackage{algorithm}
\usepackage{array}
\usepackage[caption=false,font=normalsize,labelfont=sf,textfont=sf]{subfig}
\usepackage{textcomp}
\usepackage{stfloats}
\usepackage{url}
\usepackage{verbatim}
\usepackage{graphicx}
\usepackage{cite}
\usepackage{multirow}
\usepackage{CJKutf8}  
\usepackage{booktabs}
\usepackage{tabularx}
\usepackage{caption}
\usepackage{pifont}  
\usepackage{xltabular}
\usepackage{makecell}   
\usepackage{enumitem}
\usepackage{colortbl} 
\usepackage[colorlinks,linkcolor=red,anchorcolor=blue,citecolor=green]{hyperref}

\definecolor{lightgray}{gray}{0.9} 

\newcolumntype{C}[1]{>{\centering\arraybackslash}p{#1}} 
\newcolumntype{L}{>{\arraybackslash}X} 
\newcolumntype{Y}{>{\centering\arraybackslash}X} 
\newcolumntype{H}[1]{>{\centering\arraybackslash}m{#1}}

\begin{document}
\bstctlcite{IEEEexample:BSTcontrol}

\title{4D Driving Scene Generation With Stereo Forcing}


\author{Hao Lu$*$, Zhuang Ma$*$, Guangfeng Jiang, Wenhang Ge, Bohan Li,  \\
Yuzhan Cai, Wenzhao Zheng, Yunpeng Zhang, Yingcong Chen$\dagger$
\thanks{
Hao Lu, Wenhang Ge, Yingcong Chen are with the Hong Kong University of Science and Technology (Guangzhou), Guangzhou, China (e-mail: hlu585@connect.hkust-gz.edu.cn; gewenhang01@gmail.com; yingcongchen@ust.hk).

Guangfeng Jiang is with the University of Science and Technology of China (e-mail: jgf1998@mail.ustc.edu.cn).

Bohan Li is with Shanghai Jiao Tong University, Shanghai, China (e-mail: bohan.li@sjtu.edu.cn).

Wenzhao Zheng is with the University of California, Berkeley (e-mail: wenzhao@berkeley.edu).


Hao Lu and Zhuang Ma contributed equally to this work.

Corresponding author: Yingcong Chen.
}
}

\markboth{Journal of \LaTeX\ Class Files,~Vol.~14, No.~8, August~2021}%
{Shell \MakeLowercase{\textit{et al.}}: Task_Assignments: VLA Survey}



\IEEEtitleabstractindextext{%
\begin{abstract}
Current generative models struggle to synthesize dynamic 4D driving scenes that simultaneously support temporal extrapolation and spatial novel view synthesis (NVS) without per-scene optimization. Bridging generation and novel view synthesis remains a major challenge. We present PhiGenesis, a unified framework for 4D scene generation that extends video generation techniques with geometric and temporal consistency. Given multi-view image sequences and camera parameters, PhiGenesis produces temporally continuous 4D Gaussian splatting representations along target 3D trajectories. In its first stage, PhiGenesis leverages a pre-trained video VAE with a novel range-view adapter to enable feed-forward 4D reconstruction from multi-view images. This architecture supports single-frame or video inputs and outputs complete 4D scenes including geometry, semantics, and motion. In the second stage, PhiGenesis introduces a geometric-guided video diffusion model, using rendered historical 4D scenes as priors to generate future views conditioned on trajectories. To address geometric exposure bias in novel views, we propose Stereo Forcing, a novel conditioning strategy that integrates geometric uncertainty during denoising. This method enhances temporal coherence by dynamically adjusting generative influence based on uncertainty-aware perturbations. Our experimental results demonstrate that our method achieves state-of-the-art performance in both appearance and geometric reconstruction, temporal generation and novel view synthesis (NVS) tasks, while simultaneously delivering competitive performance in downstream evaluations.  Homepage is at \href{https://jiangxb98.github.io/PhiGensis}{PhiGensis}.
\end{abstract}

\begin{IEEEkeywords}
4D Driving Generation, Novel View, Stereo Forcing 
\end{IEEEkeywords}}

\maketitle

\IEEEdisplaynontitleabstractindextext

\IEEEpeerreviewmaketitle

\section{Introduction}

Building robust autonomous driving systems requires accurate perception, prediction, and decision-making in dynamic environments~\cite{chen2024end,hu2023planning,ma2024vision,kong2025multi}. To develop and evaluate such systems at scale, driving simulators that can synthesize diverse, realistic, and temporally consistent 4D scenes are becoming increasingly essential. However, collecting and annotating large-scale real-world driving datasets is both expensive and time-consuming, often limited by sensor coverage, weather, occlusion, and scene diversity. To address these limitations, generative models have emerged as a promising alternative, enabling the controllable and scalable creation of synthetic driving scenes for perception and planning tasks.

In particular, diffusion-based models have achieved remarkable success in high-fidelity image and video generation, and are being actively explored for urban scene simulation. Building on this success, several works~\cite{gao2023magicdrive, wen2024panacea, wang2023drivedreamer, mao2024dreamdrive, gao2024magicdrivedit} have adapted diffusion models to urban street-view video generation for autonomous driving, conditioning generation on inputs such as text prompts, BEV maps, and object bounding boxes. While effective at producing photorealistic videos, these methods are tightly coupled to predefined ego trajectories and restricted to fixed camera perspectives. Consequently, they fail to support novel view synthesis (NVS)—a capability critical for autonomous driving planning and simulation, which requires flexibility across diverse viewpoints.

While NeRF~\cite{wu2023mars, yang2023emernerf} and 3D Gaussian Splatting (3DGS)-based~\cite{chen2023periodic, zhou2024drivinggaussian, yan2024street, chen2024omnire, huo2025egsral} methods enable novel view rendering, they inherently lack generative capability for new scene content.
To combine the advantages of both paradigms, hybrid methods~\cite{mao2024dreamdrive, gao2024magicdrive3d} like DreamDrive~\cite{mao2024dreamdrive} and MagicDrive3D~\cite{gao2024magicdrive3d} jointly model scene reconstruction and generation. Nevertheless, these methods still require per-scene optimization—severely limiting their scalability and ability to generalize across unseen driving scenes.

To overcome the limitations of per-scene reconstruction, recent works~\cite{wang2024stag, lu2024infinicube, guo2025dist, ren2025gen3c} propose frameworks that leverage depth information as a bridge, achieving both scene generation and novel view synthesis without explicit scene-specific optimization.
For instance, Stage-1~\cite{wang2024stag} and Gen3C~\cite{ren2025gen3c} utilize predicted depth to enable 4D-consistent scene synthesis. InfiniCube~\cite{lu2024infinicube} adopts a sequential multi-module design, but suffers from accumulated errors across stages.
DiST-4D~\cite{guo2025dist} aggregates LiDAR point clouds and applies diffusion-based completion to generate depth-supervised reconstructions.
However, during inference, the quality of predicted depth degrades—severely undermining the spatiotemporal consistency of generated 4D scenes.
These limitations highlight the need for a more unified and robust solution that ensures geometric fidelity throughout the generative process.

To address these limitations, we propose PhiGenesis, a unified and efficient framework for geometry-aware 4D generation that bridges video generation and 3D scene representation in an end-to-end manner. Unlike prior methods that rely on depth prediction or per-scene optimization, PhiGenesis directly generates temporally continuous 4D Gaussian splatting representations conditioned on historical observations and future 3D trajectories. This is achieved through a two-stage design that integrates the strengths of pre-trained video diffusion models with feed-forward 4D reconstruction pipelines.

In stage 1, we extend a pre-trained video variational autoencoder (VAE) to support 4D scene reconstruction by introducing a range-view adapter. This adapter fuses multi-view features from the decoder layers to produce 4D Gaussian representations capturing geometry, semantics, and motion—without any per-scene tuning. By doing so, PhiGenesis effectively enables real-time, generalist 4D reconstruction from single frames or video clips, supporting both monocular and multi-view inputs. In stage 2, PhiGenesis incorporates a geometry-guided video diffusion model that synthesizes future 4D scenes along a given trajectory, conditioned on the reconstructed history. To enforce temporal and geometric consistency in unseen views, we introduce a novel technique called Stereo Forcing, which dynamically modulates the generative denoising process using uncertainty-aware perturbations of the historical latent features. This mechanism corrects geometric inconsistencies by leveraging the implicit priors of the diffusion model while preserving the physical coherence of the underlying 3D structure.

We validate our approach across multiple challenging benchmarks for autonomous driving scene generation, including 4D reconstruction, novel view synthesis, and trajectory-conditioned simulation. Our results demonstrate that PhiGenesis achieves state-of-the-art performance across tasks, with strong generalization and scalability. Our contributions can be summarized as follows:
\begin{itemize} 


    \item We propose PhiGenesis, an end-to-end framework for 4D driving scene generation.

    
    \item We propose Stereo Forcing, a technique to enhance geometric consistency by encouraging the model to focus on depth-ambiguous regions during training. 
    
    \item Extensive experimental evaluations conducted on large-scale autonomous driving datasets confirm that PhiGenesis attains state-of-the-art performance across a diverse set of benchmarks.
\end{itemize}

\section{Related Work}

\subsection{Driving Generation}
Recently, diffusion-based methods~\cite{ho2020denoising, sohl2015deep, song2020score} have become the dominant paradigm in image and video generation. Building on this success, several works~\cite{li2024hierarchical, gao2023magicdrive, gao2025vista, gao2024magicdrivedit, li2023bridging, wang2024driving, mao2024dreamdrive, wang2024stag, wen2024panacea} have extended diffusion models to the generation and reconstruction of autonomous driving scenes.

\noindent\textbf{Video Driving Generation}. The MagicDrive~\cite{gao2023magicdrive} generates high-quality street-view videos by encoding multiple 3D geometric signals as conditional inputs and employing a cross-view attention mechanism to enhance frame-to-frame consistency. Similarly, Panacea~\cite{wen2024panacea} and DriveDreamer~\cite{wang2023drivedreamer} leverage cross-frame modeling to improve temporal coherence. However, these methods lack accurate and coherent 3D spatial representations.

\noindent\textbf{4D Driving Generation.} To address the need for 4D-aware scene generation, MagicDrive3D~\cite{gao2024magicdrive3d} combines video generation with reconstruction pipelines to synthesize 4D unbounded scenarios. InfiniCube~\cite{lu2024infinicube} introduces voxel-based intermediate representations, while UniScene~\cite{li2025uniscene} leverages occupancy grids to generate both LiDAR point clouds and video frames, achieving multimodal consistency through point cloud reprojection and diffusion-based completion. DiST-4D~\cite{guo2025dist} further improves novel view reconstruction by generating scene clouds via depth maps, though it still suffers from depth degradation during inference, limiting 4D consistency and realism.

To overcome these challenges, our PhiGenesis decodes 3D geometry using a dedicated 3DGS Adaptor and introduces Stereo Forcing to explicitly regularize depth consistency across frames. Unlike prior methods that rely on voxel or occupancy-based intermediates and suffer from depth quality degradation, PhiGenesis directly enhances the geometric coherence of dynamic scenes, enabling high-fidelity and consistent 4D generation without per-scene optimization.

\subsection{Scene Reconstruction}
3DGS~\cite{gaussian_splatting} has recently demonstrated impressive performance in novel view synthesis and real-time rendering. In autonomous driving, several methods~\cite{chen2023periodic, zhou2024drivinggaussian, yan2024street, chen2024omnire, huo2025egsral} apply 3DGS for dynamic scene reconstruction. To further improve generalization to unseen views, recent approaches~\cite{zhao2024drivedreamer4d, ni2024recondreamer, zhao2025recondreamer++, fan2024freesim} incorporate generative priors.
DriveDreamer4D~\cite{zhao2024drivedreamer4d} uses a world model to generate new trajectory videos for joint training with reconstruction models. ReconDreamer~\cite{ni2024recondreamer} builds a restoration dataset from degraded reconstructions and real sensor images to enhance novel view repair and improve synthetic-to-real consistency. ReconDreamer++~\cite{zhao2025recondreamer++} further introduces Novel Trajectory Deformation Networks (NTDNet) to bridge domain gaps through learnable spatial deformations.

Despite these advances, existing methods still rely on per-scene optimization and suffer from geometric inconsistencies in world model predictions. To overcome these limitations, we propose a generative 4D driving scene framework that synthesizes dynamic scenes without scene-specific optimization, leveraging a 3D cache mechanism to enhance both temporal and spatial coherence.

\subsection{Diffusion Forcing}
Traditional diffusion models~\cite{ho2020denoising, sohl2015deep, song2020score} apply uniform noise across all tokens. Diffusion Forcing (DF)~\cite{chen2024diffusion} introduces per-frame noise scheduling within causal state-space models, enabling flexible denoising and unifying next-token prediction with diffusion. CausVid~\cite{yin2024slow} extends DF to causal transformers, mitigating error accumulation in autoregressive video generation via causal attention, and improving efficiency and stability for long-form video synthesis. More recently, Song et al.~\cite{song2025history} propose the non-causal Diffusion Forcing Transformer (DFoT), which removes causal constraints to allow arbitrary-length history conditioning. By introducing History Guidance, DFoT enhances temporal consistency and supports robust long-sequence generation, outperforming prior approaches in handling variable-length contexts and complex temporal dynamics. 

While these advances have shown strong potential in video modeling, the idea of adaptive noise scheduling has not yet been explored for 4D scene generation. In this work, we adapt the principles of DF to the 4D setting: by injecting different levels of noise into different depth regions, we leverage the generative model’s geometric completion ability to recover consistent and accurate depth across frames, thereby improving 4D spatial-temporal coherence.



\section{Preliminary}
\label{pre}
Our PhiGensis is based on the video diffusion model, generalized Gaussian reconstruction and diffusion forcing technology.

\textbf{Video Diffusion Models (VDMs)} are a subset of generative models tailored for video generation. Their core architecture involves three components: an encoder maps raw video data $D$ to latent representations $z(0) = E(\mathcal{I} )$; a diffusion model iteratively denoises $\mathcal{N}(0, I)$ noise; and a decoder reconstructs generated videos $\hat{V} = D(z(0))$. To enhance VDM training stability and sampling speed, Rectified Flow is adopted. It defines a linear sample path between data distribution $p_0$ and noise distribution $p_1$:
\begin{equation} 
\label{eq:zaa}
z(t) = (1-t)z(0) + t\epsilon \quad (\epsilon \sim \mathcal{N}(0, I)),
\end{equation}
where $z(t)$ denotes the sample state at time step $t \in [0,1]$. A velocity prediction loss optimizes the network $\Theta$ (parameterizing velocity $v_\Theta$):
\begin{equation} 
\label{eq:bbb}
\mathcal{L}_{RF} = \mathbb{E}_{z(0),\epsilon,t} \left\| v_\Theta(z(t),t) - (z(0)-\epsilon) \right\|^2.
\end{equation}

\textbf{Generalizable Guassian Model.} Our method builds on the success of generalizable Gaussian models (GGM)~\cite{tian2024drivingforward,lu2024drivingrecon,yang2024storm}. GGM accepts a set of images and directly outputs a 3D or 4D representation of the video, especially, a set of Gaussian points $P$. Each Gaussian is represented by 14 parameters, including a center $\mathbf{u}\in\mathbb{R}^3$, a scaling factor $\mathbf{s}\in\mathbb{R}^3$, a quaternion rotation $\mathbf{q}\in\mathbb{R}^4$, an opacity $\mathbf{\alpha}\in\mathbb{R}$, and a color feature $\mathbf{c}\in\mathbb{R}^3$. GGM presents a novel opportunity for real-time scene reconstruction in autonomous driving. However, end-to-end 4D generation for driving has rarely been explored.

\textbf{Diffusion Forcing.} Traditionally, diffusion models are trained using uniform noise levels across all tokens. DF~\cite{chen2024diffusion} proposes training sequence diffusion models with independently varied noise levels per frame. Although DF provides theoretical and empirical support for this approach, their work focuses on causal, state-space models. CausVid~\cite{yin2024slow} builds on DF by scaling it to a causal transformer, creating an autoregressive video foundation model. History guided forcing~\cite{song2025historyguidedvideodiffusion} extends the flexibility of DF by developing both the theory and architecture for non-causal, state-free models, enabling new, unexplored capabilities in video generation. Self forcing~\cite{huang2025selfforcing} trains autoregressive video diffusion models by simulating the inference process during training, performing autoregressive rollout with KV caching. Although these methods alleviate the exposure bias in long video generation, they ignore spatial geometric uncertainty.

\begin{table}[!t]
\centering
\caption{Summary of Mathematical Symbols in the Manuscript}
\label{tab:math_symbols}
\resizebox{\linewidth}{!}{
\begin{tabular}{l l p{7cm}}
\toprule
\textbf{Symbol} & \textbf{Mathematical Form} & \textbf{Definition} \\
\midrule
$P$ & Set of Gaussian points & A set of 3D Gaussian points output by the Generalizable Gaussian Model (GGM), representing the core 3D/4D scene structure. \\
\addlinespace[0.5ex]
$\mathcal{I}$ & $\mathcal{I} = \{I_{v,tar} \mid v \in \mathcal{V}, t \in \mathcal{T}_{\text{obs}}\}$ & Input set of multi-view images, where $I_{v,tar}$ denotes the image from view $v$ at observation time $t$. \\
\addlinespace[0.5ex]
$\mathcal{V}$ & Index set of camera views & Set of indices for distinct camera views (e.g., front/left/right cameras in autonomous driving). \\
\addlinespace[0.5ex]
$\mathcal{T}_{\text{obs}}$ & Index set of observation times & Set of indices for time steps with observed input images (historical time steps for 4D generation). \\
$\mathcal{T}_{\text{future}}$ & Index set of future times & Set of indices for time steps where 4D scenes are generated (future time steps). \\
\addlinespace[0.5ex]
$K_v$ & Camera intrinsic matrix & Intrinsic parameters of camera $v$ (focal length, principal point, skew coefficients). \\
\addlinespace[0.5ex]
$R_{v,t}, T_{v,t}$ & Rotation matrix and translation vector (the camera extrinsics) & Time-varying extrinsics of camera $v$ at time $t$: $R_{v,t}$ (3×3 rotation) and $T_{v,t}$ (3D translation) define camera pose in world coordinates. \\
\addlinespace[0.5ex]
$I_{v,ren}$ & Rendered multi-view video frame & Image rendered from 4D Gaussian set $\mathcal{G}$ for view $v$ (used as geometric guidance). \\
\addlinespace[0.5ex]
$E$ & Pre-trained video VAE encoder & Encoder of the video VAE (maps 2D images to latent representations). \\
\addlinespace[0.5ex]
$D$ & Pre-trained video VAE decoder & Decoder of the frozen video VAE (maps latent representations to  2D images). \\
\addlinespace[0.5ex]
$z$ & $z \in \mathbb{R}^{h \times w \times c \times v \times t}$ & Latent tensor from the pre-trained video VAE encoder (dimensions: height $h$, width $w$, channel $c$, view $v$, time $t$). \\
\addlinespace[0.5ex]
$z_{v,ren}, z_{v, tar}$ & Latent vectors & $z_{v,ren} = E(I_{v,ren})$: Latent of rendered image $I_{v,ren}$; $z_{v, tar} = E(I_{v,tar})$: Latent of target image $I_{v,tar}$.  \\ 
\addlinespace[0.5ex]
$z_{v, tar}^n$ & Noisy targeted latent vectors & Noise latent of $z_{v, tar}$: $z_{v, tar}^n = (1-t) \cdot z_{v, tar} + t \cdot \epsilon$ ($\epsilon$ is standard Gaussian noise). \\
\addlinespace[0.5ex]
$\epsilon$ & Gaussian noise sample & Random noise sampled from $\mathcal{N}(0, I)$ (used to corrupt latents in diffusion training). \\
\addlinespace[0.5ex]
$\mathcal{L}_{RF}$ & Flow Matching loss & Loss for training the multi-view video diffusion model. \\
\addlinespace[0.5ex]
$t$ & Noise level  & the diffusion noise level ($t \in (0,1)$; $t=0$ = clean, $t=1$ = full noise). \\
\addlinespace[0.5ex]
\addlinespace[0.5ex]
$u$ & Uncertainty map & 2D map encoding geometric uncertainty (options: random noise, classification entropy, localization potential). \\
\addlinespace[0.5ex]
$\omega$ & Stereo forcing scale & The stereo forcing scale of weighting geometric contribution.  \\
\addlinespace[0.5ex]
$\mathcal{SF}(\cdot)$ & Stereo Forcing function & Function integrating uncertainty into latents: $\mathcal{SF}(u, z) =  u \cdot z$ (balances clean latents and uncertainty). \\
\bottomrule
\end{tabular}
}
\end{table}

\section{Method}

We present PhiGenesis, a novel framework that extends video generation techniques to 4D generation. Given a set of input images $\mathcal{I} = \{I_{v,tar} \mid v \in \mathcal{V}, t \in \mathcal{T}_{\text{obs}}\}$ captured from views $\mathcal{V}$ over observation times $\mathcal{T}_{\text{obs}}$, along with camera intrinsics $K_v$ and time-varying extrinsics $[R_{v,t} | T_{v,t}]$, our framework supports both single and multi-frame inputs. Conditioned on future 3D trajectories $\Gamma_{future} = \{\gamma_t\}_{t=0}^\mathcal{T}$, we generate a 4D scene representation $\mathcal{G} = \{G_t\}_{t=0}^\mathcal{T}$ using Gaussian Splatting. This allows for efficient, continuous rendering of dynamic scenes across time.  In Sec.~\ref{base}, we explain how to train the video generation model to generate 4D Gaussian representation, ultimately enabling 4D generation, termed as PhiGenesis. Further, PhiGenesis is improved by Stereo Forcing is proposed to solve the geometric exposure bias, which will be discussed in Sec.~\ref{SF}.

\begin{figure*}[!t]
    \centering
    \includegraphics[width=1.0\linewidth]{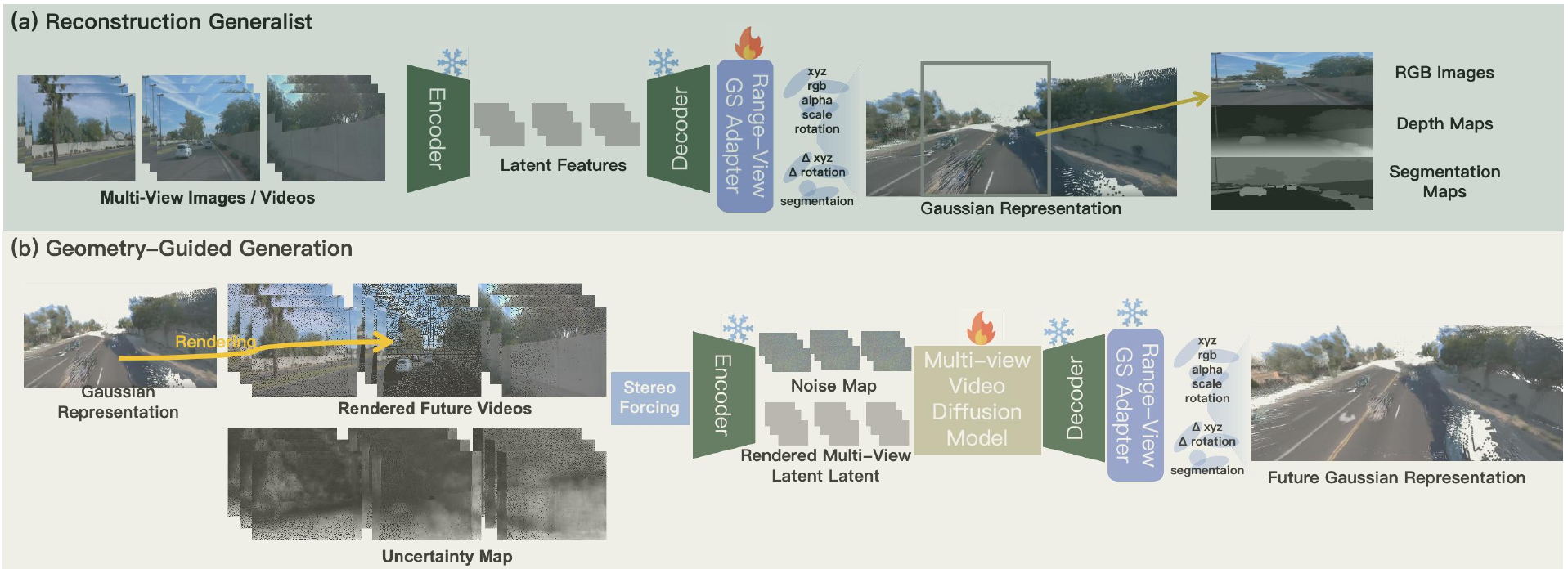}
    \caption{\textbf{Overall framework of the proposed PhiGenesis.} (1) Stage 1 aims to train a 4D reconstrucion generalist. Multi-view images are first fed into a fixed, pre-trained video VAE. The multi-scale features extracted from the decoder of the video VAE are then passed through a range-view adapter to reconstruct the complete 4D scene (including optical flow, etc.). (2) The objective of Stage 2 is to enhance geometric consistency generation. We project the 4D scenes reconstructed based on history onto the future trajectory perspective. The rendered video is denoised according to geometric uncertainty by stereo forcing and then sent to the pre-trained encoder to obtain the rendered multi-view latent. The rendered multi-view latent and noise map are fed into the multi-view video diffusion model to generate the latent of the multi-view video of the target trajectory. The latent of multi-view video is fed into the pre-trained video decoder and the GS adapter of range-view to generate the 4D scene corresponding to the target trajectory.
    }
    \label{fig_overall}
\end{figure*}

\begin{figure}[!t]
    \centering
    \includegraphics[width=1.0\linewidth]{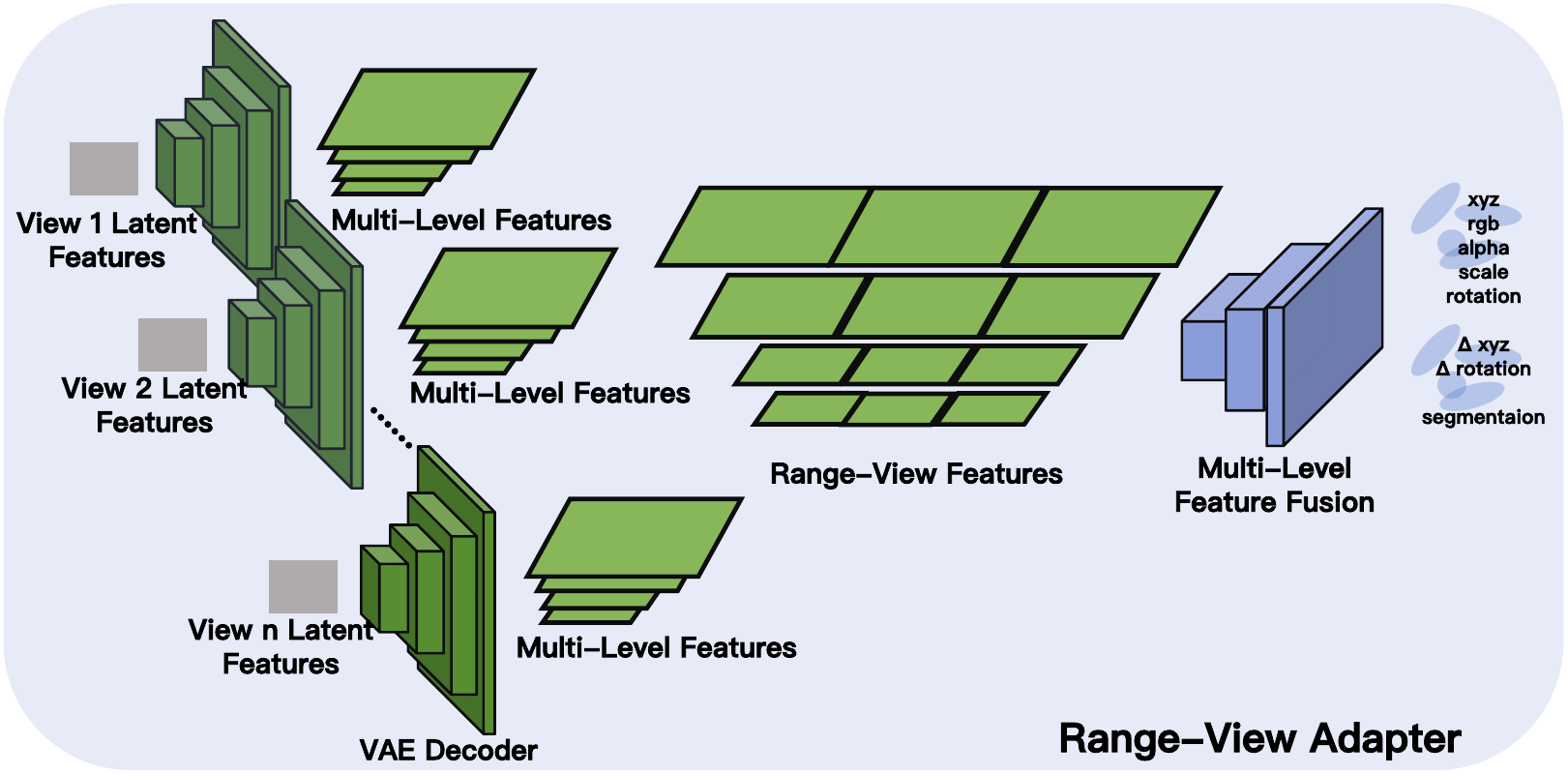}
    \caption{\textbf{The framework of the range-view adapter.} The multi-view latent is fed into the decoder of the pre-trained video. The multi-scale features of the decoder are concatenated in the form of range views and fused through a feature pyramid form to predict the Gaussian representation.
    }
    \label{fig:rg}
\end{figure}

\subsection{PhiGenesis}
\label{base}

PhiGenesis is designed to generate high-quality 4D scenes that encapsulates both the geometry and the appearance information. To achieve this, we need to solve two problems: (1) how to extend the video generation model to 4D generation, and (2) how to utilize historical geometric information to improve temporal consistency.

\subsubsection{Generalist 4D Reconstruction} Most 4D generation methods first generate some videos and then optimize per scene, which is the time consumption at the hour level~\cite{mao2024dreamdrive,zhao2025drivedreamer4d,gao2024magicdrive3d}. Fortunately, the feed-forward reconstruction method can directly convert multi-view images to 4D Gaussian representation, which provided the opportunity for real-time 4D reconstruction~\cite{yang2024storm,lu2024drivingrecon,wei2025omni}. Based on this, PhiGenesis try to extend the video VAE model to 4D reconstruction ability, which can generate 4D geometry and semantic information simultaneously.

Specifically, PhiGenesis feed single-image or video sequences ($\mathcal{I} = \{I_{v,tar} \mid v \in \mathcal{V}, t \in \mathcal{T}_{\text{obs}}\}$) captured from multi-views $\mathcal{V}$ into a fixed video VAE~\cite{kong2024hunyuanvideo} to get latent feature $e \in h \times w \times c \times v \times t$. Here, $v$ and $t$ refer to the number of perspectives and the temporal dimension after compression. It is worth noting that the pre-trained video VAE supports image or video input, adapting to diverse temporal contexts~\cite{kong2024hunyuanvideo}. Then the multi-view latent feature $e \in h \times w \times c \times v \times t$ is fed into the decoder of video VAE $D$. The native decoder reconstructs each view separately, which causing the lack of multi-view information. Therefore, the range-view representation~\cite{kong2023rethinking}, widely used in autonomous driving, is selected as the way to fuse multi-view images to predict the 4D Gaussian. 

As shown in Fig.~\ref{fig:rg}, the range-view adapter integrates features from different view at different layers, and then fused the features of different layers to obtain the final feature. The range-view adapter employs two convolutional blocks to convert features into segmentation \(\mathbf{c} \in \mathbb{R}^{C}\), depth regression \(\mathbf{d_r} \in \mathbb{R}^{1}\),  alpha \(\mathbf{a} \in \mathbb{R}^{1}\), scale \(\mathbf{r} \in \mathbb{R}^{3}\), rotation \(\mathbf{r} \in \mathbb{R}^{3}\) and optical flow \([\Delta x, \Delta y, \Delta z]\). The activation functions for RGB color, alpha, scale, and rotation are consistent with those in \cite{lu2024drivingrecon}. Through the range-view adapter, we can fully fuse multiview images and output Gaussian representations and their corresponding semantic information. During this training process, the video VAE is frozen, and only the range-view adapter is trained. During the training process, RGB rendering, depth and segmentation supervision $\mathcal{L}_{render}$ will be used. In addition, the dynamic and static separation strategy is also employed~\cite{lu2024drivingrecon}. 

\textbf{Discussion:} Many research have studied the feed-forward 4D reconstruction algorithm in driving tasks~\cite{mao2024dreamdrive,zhao2025drivedreamer4d,gao2024magicdrive3d}. Unlike them, our work focuses on extending the 4D reconstruction capabilities of the video generation models. Furthermore, the principle of our design in this part is the simplest and necessary way to integrate multi-view videos to predict 4D scenes.  There are also many papers on 4D generation in the driving field here: (1) DiST-4D only predicts the depth map simultaneously during generation rather than the complete 3D representation (specifically, the Gaussian representation)~\cite{guo2025dist}. (2) InfiniCube trains the video diffusion model and feed-forward method separately, and then converts the single-view video into a Gaussian representation in a multi-stage concatenation manner~\cite{lu2024infinicube}. 
In contrast, our approach extends the decoder's capability to directly generate the 3D Gaussian representation in an end-to-end manner.


\subsubsection{Geometric-Guided Generation} 
Based on the above steps, PhiGensis can convert the given multi-view images or videos into 4D scenes. PhiGensis further uses the existing 4D scene guidance to generate long time 4D scenes, termed \emph{Geometric-Guided Generation}. From another perspective, PhiGensis utilize the powerful painting prior of the video diffusion model to expand the existing 3D scenes to generate future scenes. 


Initially, given a set of future 3D trajectories $\Gamma_{future} = \{\gamma_t\}_{t=0}^\mathcal{T}$, we can render the multi-view video \( I_{v,ren} \) from the 4D scene representation $\mathcal{G} = \{G_t\}_{t=0}^\mathcal{T}$. Then, we employ the frozen VAE encoder \( E \) to encode the rendered multi-view video \( I_{v,ren} \) and the multi-view target video \( I_{v,tar} \), resulting in the latent representation \( z_{v,ren} = E(I_{v,ren}) \) and \( z_{v, tar} = E(I_{v,tar})\). The rendered multi-view features $z_{v,ren}$ is regarded as the geometric condition of the generative model. For training, we concatenate the noisy target video latent $z_{v,ren}^{n}$ (it began with a complete Gaussian noise) and the noisy rendered multi-view features $z_{v,ren}^{n}$ along the channel dimension, and feed these into the multi-view video diffusion model~\cite{gao2024magicdrivedit}. Here, the noisy target video latent $z_{v, tar}^{n} = (1 - t ) * z_{v, tar} + t * \epsilon$ is constructed by adding noise $\epsilon$ sampled from a Gaussian distribution, which began with a pure Gaussian noise map. The rendered multi-view features $z_{v,ren}^{sf} = \mathcal{SF}(z_{v,ren})$ is processed by stereo forcing $\mathcal{SF}$, i.e., $z_{v,ren}^{sf} = \mathcal{SF}(z_{v,ren})$. The stereo forcing will be disscussed in Sec.~\ref{SF}. For our model, we adopt the VAE architecture from HunyuanVideo \cite{kong2024hunyuanvideo} and leverage OpenSora V2\footnote{\url{https://github.com/hpcaitech/Open-Sora}} for diffusion pre-training. Additionally, the multi-view cross-attention fusion mechanism integrated into the diffusion model draws inspiration from MagicDirveV2 \cite{gao2024magicdrivedit}. Our training pipeline use flow matching loss $\mathcal{L}_{RF}$ to generate the multi-view videos. This method ensures the integration of geometric representation and latent representation, ultimately promoting the generation of consistent and visually coherent predictive 3D scenes.


\textbf{Discussion:} Many concurrent research paper also studied the 3D-Guided Generation~\cite{ren2025gen3c,chen2025geodrive}. Our method has three advanced aspects compared with these two concurrent works: (1) These methods rely on existing geometric reconstruction tools such as MonST3R~\cite{zhang2024monst3r}, while our algorithm can directly reconstruct historical 3D information. (2) Our method supports multi-view videos, while these methods only support monocular videos. 

\begin{figure}[!t]
    \centering
    \includegraphics[width=1.0\linewidth]{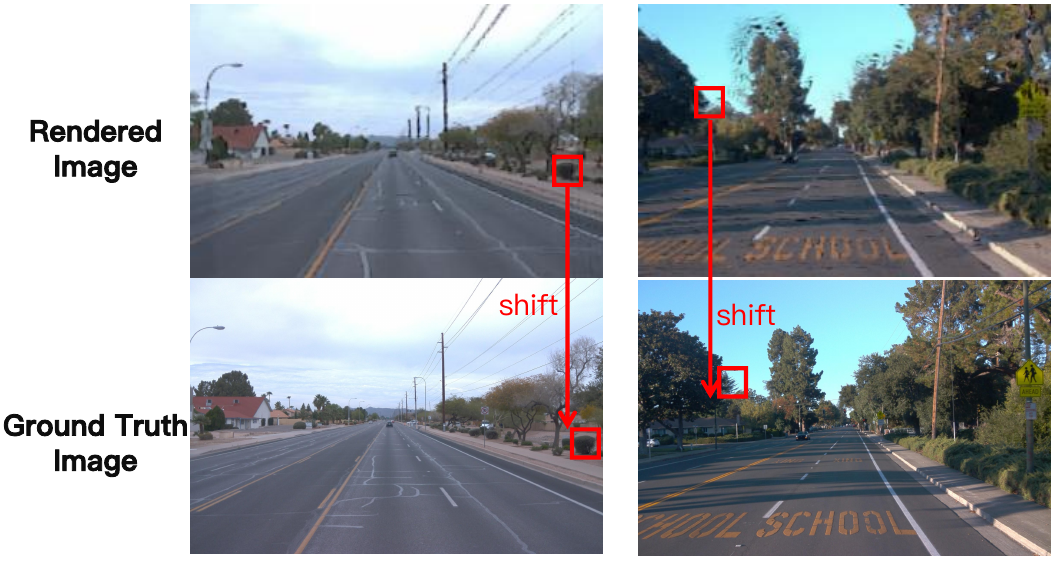}
    \caption{\textbf{The geometric exposure bias.} Due to inaccurate geometric estimation, the rendered view is not strictly accurate. Such inaccurate renderings used as geometry guidance will lead to the degradation of consistency. This requires additional information to further guide the Diffusion process to correct inconsistencies.
    }
    \label{fig:sf}
\end{figure}


\subsection{PhiGenesis with Stereo Forcing}
\label{SF}

Although geometry-guided history can effectively retains historical information, the biggest problem is that its depth is inaccurate in unseen scene, leading to poor geometric consistency in the generated 4D scene, termed the geometric exposure bias as shown in Fig.~\ref{fig:sf}. Inspired by DF~\cite{chen2024diffusion,song2025history}, we attempt to explore a 4D geometrical relationship of the denoising process. To achieve this, we propose \emph{Stereo Forcing}, which enhances the quality of 4D generation while maintaining consistency. The core idea is to let the denosing process know the geometry uncertainty, and then fix the inaccurate parts of the projection based on the generation prior. Specifically, the score is given by:
\begin{equation} 
\label{eq:UAH}
\nabla p(z_{v, tar}) + \omega \big[\nabla p\big(z_{v, tar} | \mathcal{SF}(u, z_{v,ren})\big) - \nabla p(z_{v, tar})\big],
\end{equation} where $u \in (0, 1)$ is uncertainty map. This approach differs from conventional CFG and HG~\cite{song2025history} in two ways: (1) The condition $z_{v,ren}^{k}$ is rendered from the 4D Gaussian, which retain geometric information to alleviate generative degradation. (2) Stereo forcing function $\mathcal{SF}$ introduces additional uncertain information. The further question is how to determine the geometric uncertainty  to determine stereo forcing function $\mathcal{SF}$. Here, we have attempted three common uncertain maps $u$: random noise, entropy of depth classification, and the localization potential~\cite{Park2022TimeWT}. 

The localization potential comprehensively and theoretically analyzes and quantitatively presents the uncertainties of temporal fusion, multi-view fusion, and the camera parameters of the camera on geometric estimation~\cite{Park2022TimeWT}. Inspired by his outstanding effect in perception, we also applied this uncertain map to the denosing process of 4D generation. The uncertain indicator can also be added to stereo forcing function $\mathcal{SF}(u, z_{v, ren}) = u \cdot  z_{v, ren} $. We emphasize that the form of uncertainty and the form of stereo forcing can be further enhanced, and here we only choose the most concise way to prove the effectiveness of our method.

\begin{table*}[ht]
    \centering
    \caption{\textbf{Model Training Stages.} For a fair comparison, we conducted comparisons using different resolutions on different datasets. The resolutions of R1 at waymo and nuScene are 160 × 240 and 224 × 400 respectively. The resolutions of R2 at waymo and nuScene are 576 × 1024 and 424 × 800 respectively.}
    \resizebox{\linewidth}{!}{
    \begin{tabular}{@{}lccc@{}}
        \toprule
        \textbf{Training Configuration} & \textbf{Stage 1} & \textbf{Stage 2.1} & \textbf{Stage 2.2} \\ 
        Resolution & $R_{1}, R_{2}$ & $R_{1}$ & $R_{2}$ \\  
        Fine-Tuned Model & Range-View Adapter & Multi-View Video Diffusion & Multi-View Video Diffusion \\  
        Constraint Loss & $\mathcal{L}_{render}$ & $\mathcal{L}_{RF}$ & $\mathcal{L}_{RF}$ \\ 
        Training Steps & 10,000 & 50,000 & 50,000 \\
        Learning Rate & $2 \times 10^{-4}$ & $1 \times 10^{-4}$ & $5 \times 10^{-5}$ \\  
        LR Scheduler & Cosine Annealing & Cosine Annealing with Restarts & Cosine Annealing with Restarts \\  
        Weight Decay & $1 \times 10^{-4}$ & $5 \times 10^{-5}$ & $5 \times 10^{-5}$ \\ 
        Gradient Norm Clip & 1.0 & 1.0 & 1.0 \\  
        Optimizer & AdamW & AdamW & AdamW \\ \bottomrule
    \end{tabular}
    \label{tab:training}
    }
\end{table*}

\subsection{Training and Reasoning}

\textbf{Training.} Generating high-quality 4D scenes is non-trival. We adopted a two-stage training strategy: (1) Building 4D reconstruction capability. (2) Building Generative capacity. It is worth mentioning that to achieve high-quality 4D generation, we adopted a mixed-resolution training process \cite{gao2024magicdrivedit}. For the construction of 4D reconstruction capabilities, we have implemented a hybrid strategy of high-resolution single-frame and low-resolution multi-frame. For the construction of generation capabilities, we first adopt a process from low resolution to high resolution to accelerate convergence~\cite{gao2024magicdrivedit}. The details of the training stage are shown in Tab.~\ref{tab:training}.

\textbf{Inference.} Our model supports single-frame or multi-frame input to generate 4D scenes. Specifically, both single-frame and multi-frame videos can be reconstructed into 3D scenes, and geometric conditions can be rendered based on future trajectories as guidance. 
Moreover, the generated 4D scene can be used to predict and render subsequent geometric states, enabling a rollout-style generation process.



\begin{table*}[tb]
    \centering
    \caption{\textbf{\textbf{Comparison with state-of-the-art methods on the Waymo and nuScenes datasets.} Metrics reported include Peak Signal-to-Noise Ratio ( PSNR), Structural Similarity Index Measure (SSIM), Depth RMSE (D-RMSE), Learned Perceptual Image Patch Similarity (LPIPS), and Pearson Correlation Coefficient (PCC). Higher values are better for PSNR, SSIM, and PCC; lower values are better for D-RMSE and LPIPS.}}
    \resizebox{0.8\textwidth}{!}{
    \begin{tabular}{l|ccc|cccc}
    \toprule
    \multirow{2}{*}{\textbf{Method}} & \multicolumn{3}{c|}{\textbf{Waymo}} & \multicolumn{4}{c}{\textbf{nuScenes}} \\
    & PSNR$\uparrow$ & SSIM$\uparrow$ & D-RMSE$\downarrow$ 
    & PSNR$\uparrow$ & SSIM$\uparrow$ & LPIPS$\downarrow$ & PCC$\uparrow$ \\
    \midrule
    \multicolumn{8}{l}{\textit{Per-Scene Optimization Methods}} \\
    EmerNeRF~\cite{yang2023emernerf} & 24.51 & 0.738 & 33.99 & 18.45 & 0.582 & 0.502 & 0.061 \\
    3DGS~\cite{kerbl20233d} & 25.13 & 0.741 & 19.68 & 19.67 & 0.603 & 0.436 & 0.094 \\
    PVG~\cite{Chen2023PeriodicVG} & 22.38 & 0.661 & 13.01 & 18.98 & 0.567 & 0.481 & 0.072 \\
    DeformableGS~\cite{yang2023deformable} & 25.29 & 0.761 & 14.79 & 20.12 & 0.622 & 0.422 & 0.105 \\
    \midrule
    \multicolumn{8}{l}{\textit{Generalizable Feed-Forward Methods}} \\
    LGM~\cite{tang2024lgm} & 23.59 & 0.691 & 8.02 & 22.15 & 0.672 & 0.318 & 0.342 \\
    GS-LRM~\cite{zhang2024gs} & 25.18 & 0.753 & 7.94 & 23.41 & 0.703 & 0.273 & 0.598 \\
    pixelSplat~\cite{pixelsplat} & 22.65 & 0.684 & 11.03 & 21.51 & 0.616 & 0.372 & 0.001 \\
    MVSplat~\cite{chen2025mvsplat} & 23.42 & 0.701 & 9.88 & 21.61 & 0.658 & 0.295 & 0.181 \\
    SCube~\cite{ren2024scube} & 25.72 & 0.783 & 5.62 & 23.85 & 0.721 & 0.258 & 0.651 \\
    DrivingForward~\cite{tian2024drivingforward} & 26.32 & 0.774 & 5.79 & 24.32 & 0.732 & 0.229 & 0.766 \\
    Omni-Scene~\cite{wei2025omni} & 26.46 & 0.786 & 5.66 & 24.27 & 0.736 & 0.237 & 0.804 \\
    STORM~\cite{yang2024storm} & 26.38 & 0.794 & 5.48 & 24.56 & 0.752 & 0.217 & 0.788 \\
    \midrule
    \textbf{PhiGensis} & \textbf{27.52} & \textbf{0.833} & \textbf{5.14} & \textbf{25.92} & \textbf{0.801} & \textbf{0.189} & \textbf{0.847} \\
    \bottomrule
    \end{tabular}
    }
    \label{tab:main_table}
\end{table*}

\begin{figure*}[!ht]
    \centering
    \includegraphics[width=0.85\linewidth]{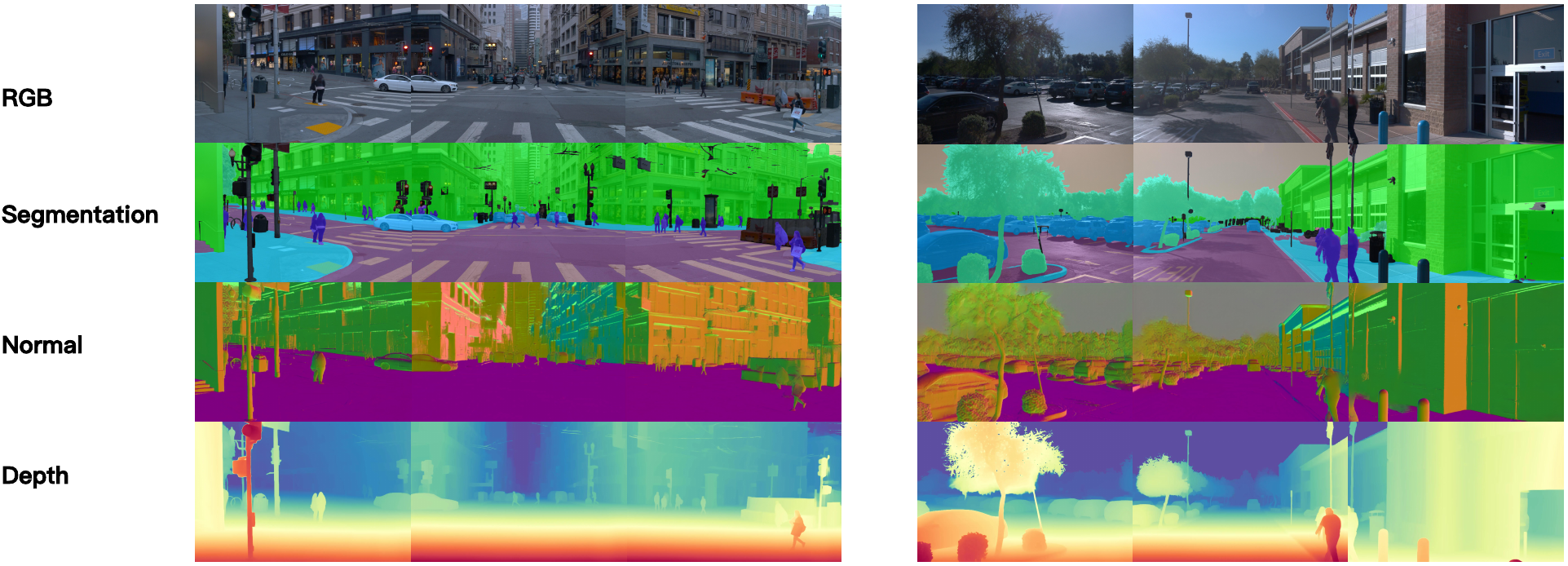}
    \caption{\textbf{Qualitative comparison on reconstruction ability.} We visualized the reconstructed RGB images, segmentation, normal and depth maps, which demonstrated extremely high quality.
    }
    \label{recon}
\end{figure*}

\section{Dataset} 

Our experimental evaluations were performed on two benchmark datasets: the Open Waymo Dataset (hereafter referred to as OWD)~\cite{sun2020scalability} and the nuScenes dataset~\cite{caesar2020nuscenes}. For OWD, we utilized its dedicated training set for model training and its validation set for performance testing, following standard evaluation protocols. Regarding the nuScenes dataset, we adjusted the annotation frequency of keyframes—originally provided at 2Hz—to 12Hz through interpolation, consistent with the approach adopted in prior research~\cite{wang2022asap,gao2023magicdrive}. We strictly followed the official data partitioning scheme for nuScenes, using 700 videos as the training corpus and 150 videos for validation.
For the generation of semantic map ground truth, we first retrained the SegFormer model~\cite{xie2021segformer} and then applied the fine-tuned model to infer pseudo semantic maps for each individual frame. To acquire dense depth ground truth, we aggregated LiDAR point clouds of static scene components across the entire scene, followed by projecting these temporally fused point clouds onto each frame; this process yielded sparse LiDAR depth maps. In line with conventional depth completion methodologies, we employed the DepthLab tool~\cite{liu2024depthlab} to densify these sparse depth maps.



\section{Experiment}

It is difficult to directly and quantitatively evaluate the ability of 4D generation without the 4D label. In this section, we attempt to evaluate our algorithm from two aspects: (1) Is it efficient to endow the video pre-trained VAE with 4D reconstruction capabilities? (2) Can this method generate higher-quality scenes with controllable trajectories?

\begin{table}[!t]
\centering
\caption{\textbf{Quantitative Evaluation of Generated Depth Maps}. We compare our method with state-of-the-art surround depth estimation approaches \cite{wei2023surrounddepth, guo2025dist} on the Waymo dataset. }
\label{tab:depth_generation_results}
\resizebox{0.99\linewidth}{!}{
\begin{tabular}{l|cccc}
\toprule
\textbf{Method} & Abs. Rel. $\downarrow$ & RMSE $\downarrow$ & $\delta < 1.25$ $\uparrow$ & $\delta < 1.25^2$ $\uparrow$ \\
\midrule
SD~\cite{wei2023surrounddepth} & 0.25 / 0.26 & 5.98 / 6.06 & 0.72 / 0.68 & 0.89 / 0.87 \\
DiST-4D~\cite{guo2025dist} & 0.18 / 0.23 & 4.90 / 5.13 & 0.82 / 0.79 & 0.93 / 0.92 \\
\textbf{PhiGensis (Ours)} & \textbf{0.16} / \textbf{0.19} & \textbf{4.58} / \textbf{4.81} & \textbf{0.86} / \textbf{0.82} & \textbf{0.95} / \textbf{0.96} \\
\bottomrule
\end{tabular}}
\end{table}

\begin{table}[!t]
\begin{center}
\caption{\textbf{Quantitative comparison of novel view synthesis}. We evaluate FID and FVD across different viewpoint shifts (±1m, ±2m, ±4m).
}
\scriptsize
\renewcommand\tabcolsep{1.0pt}
\resizebox{1.0\linewidth}{!}{%
\begin{tabular}{l|cc|cc|cc}
\toprule 
\multicolumn{1}{l|}{\multirow{2}{*}{Method}} & \multicolumn{2}{c|}{Shift $\pm\$1m$} & \multicolumn{2}{c|}{Shift $\pm\$2m$} & \multicolumn{2}{c}{Shift $\pm\$4m$} \\
\cmidrule(lr){2-3} \cmidrule(lr){4-5} \cmidrule(lr){6-7}
 & ~ FID $\downarrow$ ~ & ~FVD $\downarrow$ ~ & ~ FID $\downarrow$ ~ & ~ FVD $\downarrow$ ~ & ~ FID $\downarrow$ ~ & ~ FVD $\downarrow$ ~ \\
\midrule
PVG~\cite{chen2023PVG}      &  48.15 & 246.74  & 60.44 & 356.23 & 84.50 & 501.16  \\
EmerNeRF~\cite{yang2023emernerf}   &  37.57  &  171.47  & 52.03  & 294.55 & 76.11 &  497.85 \\ 
StreetGaussian~\cite{yan2024street} & 32.12  & 153.45  &  43.24 & 256.91 & 67.44 & 429.98 \\
OmniRe~\cite{chen2024omnire}   & 31.48  & 152.01  & 43.31  & 254.52 & 67.36 & 428.20 \\
FreeVS~\cite{wang2024freevs}   &  51.26  &  431.99  &  62.04  &  497.37  &  77.14  &  556.14 \\
DiST-4D   &  10.12 &  45.14 &  12.97  &  68.80  &  17.57  &  105.29  \\
\midrule
\textbf{Ours}   &  \textbf{9.80} &  \textbf{43.80} &  \textbf{11.71}  &  \textbf{67.54}  &  \textbf{15.51} &  \textbf{103.13}  \\
\bottomrule
\end{tabular}
}
\label{tab_NVS}
\end{center}
\end{table}

\begin{figure}[!t]
    \centering
    \includegraphics[width=0.9\linewidth]{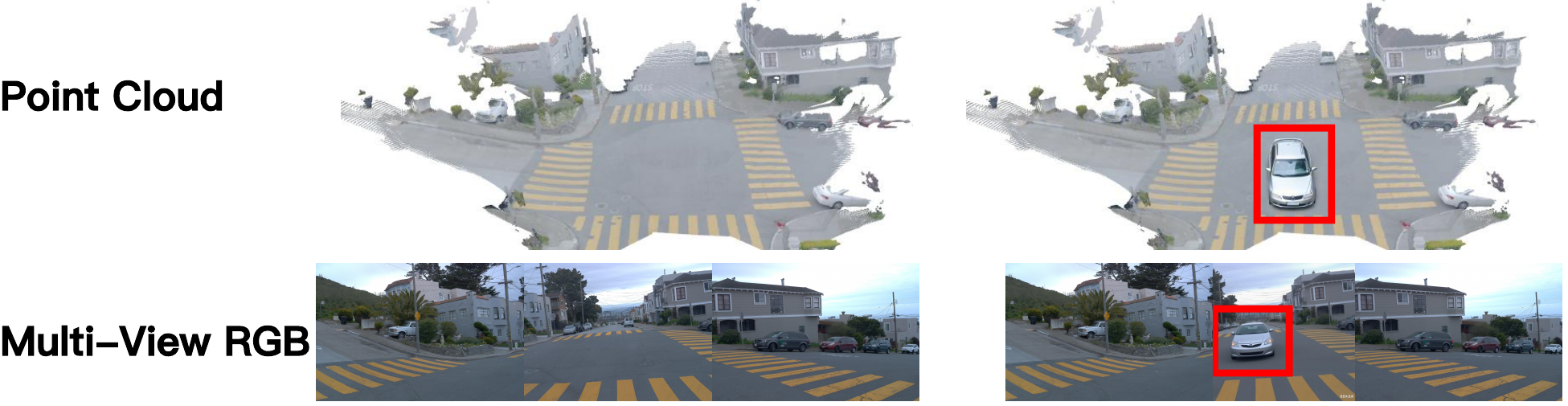}
    \caption{\textbf{Scene editing.} Various 3D assets can be inserted into the generated scene.
    }
    \label{fig:insert}
\end{figure}

\begin{figure*}[!t]
    \centering
    \includegraphics[width=0.85\linewidth]{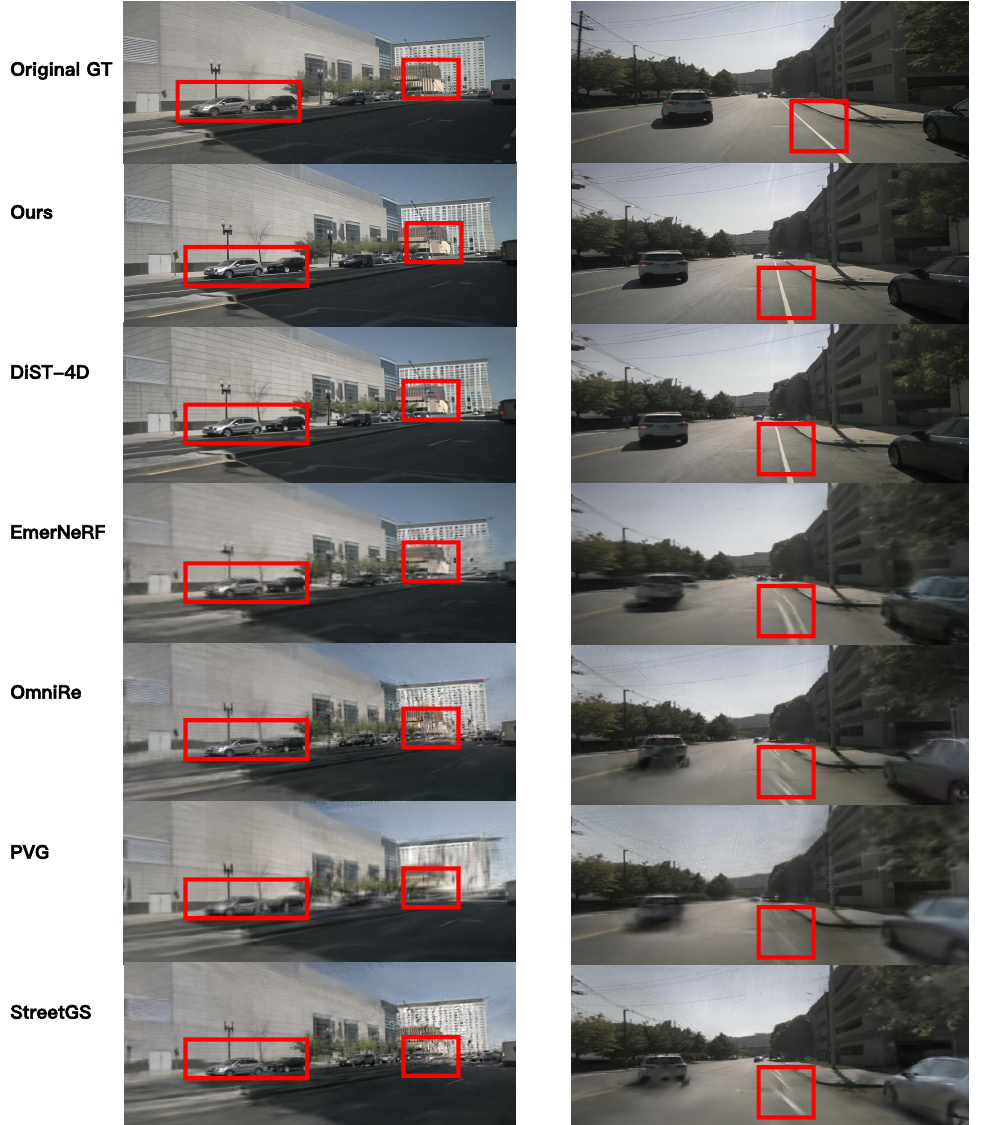}
    \caption{\textbf{Qualitative comparison on spatial NVS.} Comparison between Ours and state-of-the-art methods on spatial NVS. Please zoom in for the best observation. The red part indicates that our method has outstanding advantages in reconstructing object details, lane lines and distant views.
    }
    \label{nvs}
\end{figure*}
\textbf{Apparent reconstruction ability.} We verified the 4D reconstruction performance through two dimensions: (1) The model’s ability to reconstruct input data after the first training stage; (2) Its reconstructed generation capability following the second training stage. For the first stage, we compared our method against two broad categories of baselines: per-scene optimization methods and generalizable feed-forward models. For per-scene optimization, we evaluated against both NeRF-based and 3DGS-based approaches, including EmerNeRF~\cite{yang2023emernerf}, 3DGS~\cite{kerbl20233d}, PVG~\cite{Chen2023PeriodicVG}, and DeformableGS~\cite{yang2023deformable}. Since LiDAR data is unavailable during testing in our experimental setup, these baselines were also run without LiDAR supervision to ensure a fair comparison. For the generalizable feed-forward category, we compared with state-of-the-art large-scale reconstruction models, such as LGM~\cite{tang2024lgm}, GS-LRM~\cite{zhang2024gs}, SCube~\cite{ren2024scube}, DrivingForward~\cite{tian2024drivingforward}, and STORM~\cite{yang2024storm}. The rendered image resolutions were set to 160×240 for the Waymo dataset and 224×400 for the nuScenes dataset, consistent with the settings in STORM~\cite{yang2024storm} and OmniScene~\cite{wei2025omni}.

Quantitative results are presented in Table \ref{tab:main_table}. On the Waymo dataset, our method (“PhiGensis”) achieves a PSNR of 27.52, which is 0.86 higher than STORM (26.38) and 0.46 higher than Omni-Scene (26.46)—the two top-performing feed-forward baselines. In terms of depth accuracy (D-RMSE), our method’s score of 5.14 is 0.34 lower than STORM’s 5.48, reflecting more precise geometric reconstruction.
On the nuScenes dataset, PhiGensis maintains its lead: its PSNR of 25.92 is 1.36 higher than STORM (24.56), and its LPIPS score of 0.189 is 0.028 lower than STORM’s 0.217—demonstrating superior visual fidelity. A key observation is that generalizable feed-forward models achieve performance comparable to per-scene optimization methods in terms of photorealism, geometric accuracy, and inference speed—both in dynamic regions and full images. Notably, our method outperforms other generalizable feed-forward models in modeling scene dynamics and processing multi-timestep, multi-view data. This superior reconstruction performance can be attributed to our utilization of a pre-trained video VAE, which provides robust spatiotemporal feature priors that simpler feed-forward architectures lack. With its outstanding reconstruction performance, we can insert  objects into scenes, as shown in the Fig.~\ref{fig:insert}.

\begin{figure*}[!t]
    \centering
    \includegraphics[width=0.95\linewidth]{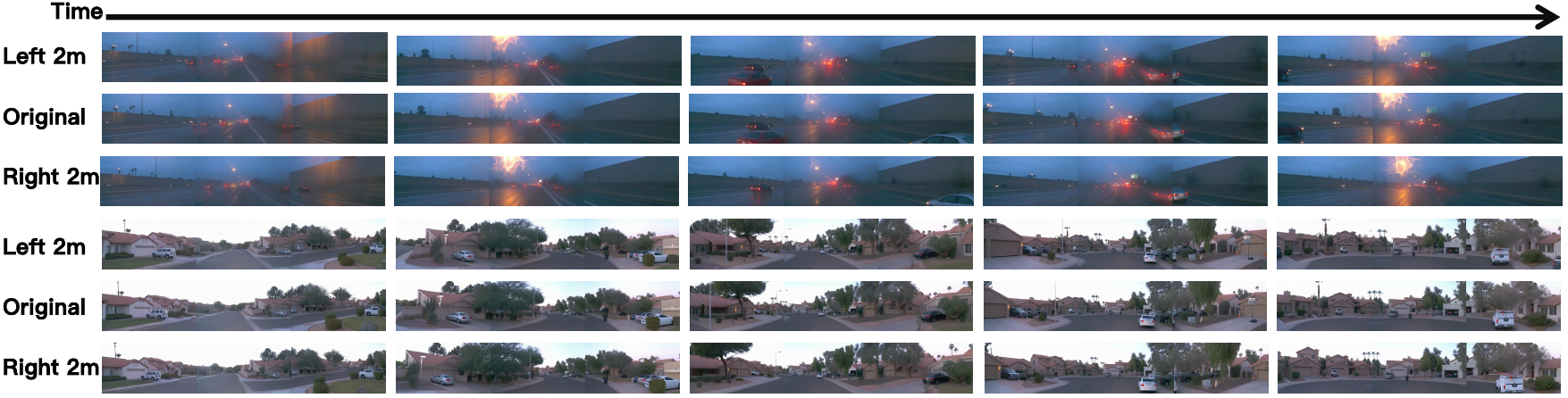}
    \caption{\textbf{New view synthesis for long videos.} In waymo's dataset, we visualized the three views generated from long-time series videos. Under long video generation, our method still maintains high quality after shifting the observation view (please zoom in for the best view).
    }
    \label{fig:demo}
\end{figure*}

\textbf{Geometry reconstruction ability.}  Following the evaluation protocol in \cite{guo2025dist}, we further report the geometric prediction performance on the Waymo dataset. As shown in Table \ref{tab:depth_generation_results}, the depth maps generated by our method (“PhiGensis (Ours)”) achieve performance comparable to—and in most cases superior to—state-of-the-art surround depth estimation approaches like SurroundDepth \cite{wei2023surrounddepth} and DiST-4D \cite{guo2025dist}. For the Absolute Relative Error (Abs. Rel.), our method achieves 0.16/0.19, which is 0.02/0.04 lower than DiST-4D (0.18/0.23) and 0.09/0.07 lower than SurroundDepth (0.25/0.26), indicating smaller erros between predicted and ground-truth depth. This confirms that the 4D generalist model trained in our first stage possesses excellent geometric reconstruction ability. As visualized in Figure \ref{recon}, our model produces clear, geometrically consistent reconstructions, further validating its strong reconstruction capability.

\textbf{New view synthesis ability.} Adopting the evaluation methodology from \cite{wang2024freevs, guo2025dist}, we focus on assessing our model’s performance on novel trajectories using Fréchet Inception Distance (FID) and Fréchet Video Distance (FVD). Specifically, we apply lateral offsets of $\tau \in \{\pm 1m, \pm 2m, \pm 4m \}$ to the camera viewpoint and compute FID/FVD between the synthesized RGB images of the novel trajectory and the ground-truth images of the original trajectory. Table \ref{tab_NVS} compares the NVS results of our method with existing approaches under these shifted viewpoints.

At a  $\tau \in \{\pm 1m\}$ shift (small viewpoint change), our FID is 9.80 and FVD is 43.80—slightly outperforming DiST-4D (10.12/45.14) and far surpassing traditional per-scene methods like PVG (48.15/246.74) and EmerNeRF (37.57/171.47) by a large margin. For the larger $\tau \in \{\pm 4m\}$ shift (more challenging viewpoint), our FID of 15.51 and FVD of 103.13 remain the lowest among all methods, outperforming DiST-4D (17.57/105.29) and exceeding StreetGaussian (67.44/429.98) by over 50 points in FID and 300 points in FVD. The quantitative results demonstrate that our method achieves substantial improvements in both FID and FVD metrics. When using conditions generated from real images as inputs, ours achieves several times better performance in these metrics, highlighting its strong capability in NVS tasks based on real data. As illustrated in Figure \ref{nvs}, our proposed method significantly outperforms previous reconstruction-based models in synthesizing novel spatial viewpoints for RGB images—with results nearly free of visual degradation and artifacts like blurring or geometric distortion.

For the second training stage, we adopt the experimental setup from SCube~\cite{ren2024scube, lu2024infinicube} to assess reconstruction quality: given 3 front-view inputs from frame $T$, we synthesize novel views at frames $T+5$ and $T+10$, then evaluate performance using PSNR, SSIM, and LPIPS. Quantitative results are presented in Table \ref{table:recon_comparison}, where our method (“PhiGensis”) outperforms all baseline approaches across all metrics and both time steps.  Notably, our method not only leverages the prior knowledge generated during training but also benefits from the “Stereo Forcing” mechanism, which contributes to synthesizing higher-quality videos with stronger temporal consistency—explaining its consistent superiority over baselines, especially at the more distant frame $T+10$.

\begin{table}[t]
\centering
\caption{\textbf{Quantitative Comparison of Cross-Temporal Rendering on the Waymo Dataset.} Metrics are calculated for synthesized views at frames $T+5$ and $T+10$ given frame $T$ as input. Higher values are better for PSNR and SSIM; lower values are better for LPIPS.}
\label{table:recon_comparison}
\resizebox{0.45\textwidth}{!}{
\begin{tabular}{lcccccc}
\toprule
 & \multicolumn{3}{c}{$T + 5$} & \multicolumn{3}{c}{$T + 10$} \\
\cmidrule(lr){2-4} 
\cmidrule(lr){5-7} 
 & PSNR$\uparrow$ & SSIM$\uparrow$ & LPIPS$\downarrow$ & PSNR$\uparrow$ & SSIM$\uparrow$ & LPIPS$\downarrow$ \\
\midrule
PixelNeRF~\cite{yu2021pixelnerf}  & 15.21 & 0.52 & 0.64 & 14.61 & 0.49 & 0.66 \\
PixelSplat~\cite{pixelsplat}   & 20.11 & 0.70 & 0.60 & 18.77 & 0.66 & 0.62 \\
DUSt3R~\cite{wang2023dust3r}  & 17.08 & 0.62 & 0.56 & 16.08 & 0.58 & 0.60 \\
MVSplat~\cite{chen2025mvsplat}   & 20.14 & 0.71 & 0.48 & 18.78 & 0.69 & 0.52 \\
MVSGaussian~\cite{liu2024mvsgaussian}  & 16.49 & 0.70 & 0.60 & 16.42 & 0.60 & 0.59 \\ 
SCube~\cite{ren2024scube}   & 19.90 & 0.72 & 0.47 & 18.78 & 0.70 & 0.49  \\ 
Infinicube~\cite{lu2024infinicube} & 20.80 & 0.73 & 0.42 & 19.93 & 0.72 & 0.45 \\
\midrule
PhiGensis & \textbf{21.41} & \textbf{0.75} & \textbf{0.38} & \textbf{20.12} & \textbf{0.74} & \textbf{0.42} \\
\bottomrule
\end{tabular}
}
\end{table}

\textbf{Generation ability.} To assess the performance of our model in long video generation, we follow the experimental paradigm proposed in \cite{lu2024infinicube}: we take the initial frames of test sequences from the Waymo Open Dataset as inputs to various video generation models, and compute the FID for generated video frames at different temporal indices. This metric is employed to quantify the degradation of video quality over extended generation horizons.

\begin{table}[h]
\centering
\caption{FID Values of Different Methods Across Various Video Frame Lengths. Lower FID values indicate higher consistency between generated videos and real-world visual distributions.}
\begin{tabular}{lcccc}
\toprule
\textbf{Method} & \textbf{FID@50} & \textbf{FID@100} & \textbf{FID@150} & \textbf{FID@200} \\
\midrule
Vista~\cite{gao2025vista}    & 130.2 & 160.4 & 195.1 & 224.8 \\
Panacea~\cite{wen2023panacea}  & 109.7 & 140.3 & 169.8 & 201.5 \\
InfiniCube~\cite{lu2024infinicube}     &  85.5 &  95.1 & 105.3 & 115.7 \\
Dist-4D~\cite{guo2025dist}     &  72.3 &  84.2 & 98.3 & 102.7 \\
\midrule
Ours wo SF &  60.4 &  73.8 & 92.6 & 91.2 \\
Ours &  55.2 &  68.6 & 82.8 & 88.3 \\
\bottomrule
\end{tabular}
\label{tab:fid_comparison}
\end{table}

As presented in Table \ref{tab:fid_comparison}, our full model (“Ours”) consistently achieves the lowest FID scores across all evaluated frame lengths (FID@50 to FID@200), demonstrating superior and sustained visual quality over long generation horizons. In contrast, baseline methods exhibit pronounced quality degradation as the video length extends beyond 100 frames: for instance, Vista’s FID increases by 72.5 \% from 130.2 (FID@50) to 224.8 (FID@200), while Panacea’s FID rises by 83.6\% over the same range. Even state-of-the-art methods like InfiniCube and Dist-4D show a 35.3\% and 42.1\% FID increase, respectively, from FID@50 to FID@200. This superior performance underscores the efficacy of our world-guided video generation strategy and guidance buffer design, which effectively mitigates the cumulative autoregressive errors that typically plague long video generation tasks. Among the buffer components, the semantic buffer plays a pivotal role in preserving high-level video quality, while the coordinate buffer resolves fine-grained ambiguities arising from motion-induced scene transformations.

\begin{figure*}[!t]
    \centering
    \includegraphics[width=0.8\linewidth]{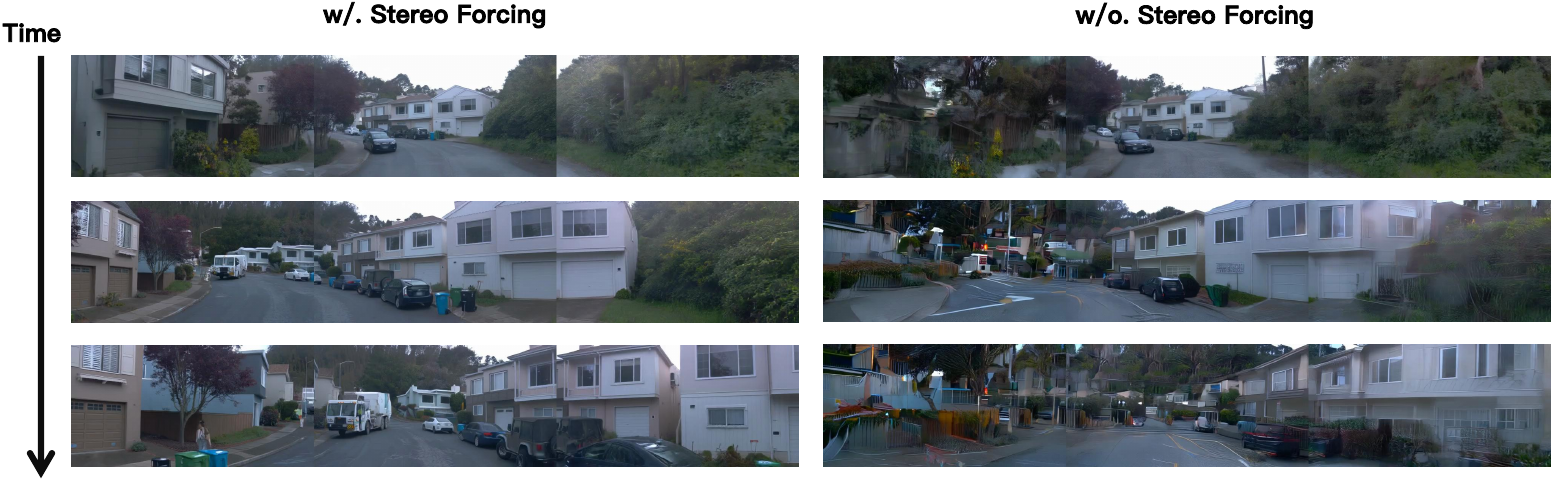}
    \caption{\textbf{Qualitative ablation on stereo forcing.} The stereo forcing can effectively reduce generative degradation and improve geometric consistency.
    }
    \label{fig:sfab}
\end{figure*}

\begin{table*}[h]
\centering
\caption{Ablation Study on Different Uncertainty Metrics for Stereo Forcing. Lower FID values indicate better video generation quality and consistency.}
\begin{tabular}{lcccc}
\toprule
\textbf{Uncertainty Metric} & \textbf{FID@50} & \textbf{FID@100} & \textbf{FID@150} & \textbf{FID@200} \\
\midrule
Baseline & 60.4 & 73.8 & 92.6 & 91.2 \\
Random Noise & 58.7 & 70.2 & 89.5 & 90.1 \\
Entropy & 57.1 & 69.3 & 86.4 & 89.2 \\
Localization Potential \cite{Park2022TimeWT} & 55.2 & 68.6 & 82.8 & 88.3 \\
\bottomrule
\end{tabular}
\label{tab_sf}
\end{table*}

\begin{table}[!t]
\begin{center}
\scriptsize
\renewcommand\tabcolsep{1.0pt}
\caption{\textbf{Quantitative Evaluation of Video Generation for Perception and Planning}. We use UniAD \cite{hu2023planning} to assess object detection, BEV segmentation, and L2 open-loop planning errors on generated videos.}
\resizebox{0.99\linewidth}{!}{
\begin{tabular}{l|cc|cccc|ccc}
\toprule 
\multicolumn{1}{l|}{\multirow{2}{*}{Method}} & \multicolumn{2}{c|}{Detection $\uparrow$} & \multicolumn{4}{c|}{BEV Segmentation $\uparrow$} & \multicolumn{3}{c}{L2 $\downarrow$}  \\
\cmidrule(lr){2-3} \cmidrule(lr){4-7} \cmidrule(lr){8-10} 
 & ~ NDS ~ & ~ mAP ~ & ~ Lan.~ & ~ Dri. ~ & ~ Div. ~ & ~ Cro. ~ & ~ 1.0 ~ & ~ 2.0 ~ &  ~ 3.0 ~  \\
\midrule
Ori. GT  & 49.85 & 37.98 & 31.31 & 69.14 & 25.93 & 14.36 & 0.51 & 0.98 & 1.65  \\
\midrule
MD~\cite{gao2023magicdrive}  & 28.36 & 12.92 & 21.95 & 51.46 & 17.10 & 5.25 & 0.57 & 1.14 & 1.95  \\
DA~\cite{yang2024drivearena}  & 30.03 & \underline{16.06} & 26.14 & 59.37 & 20.79 & 8.92 & \underline{0.56} & \underline{1.10} & \underline{1.89}   \\
Dist-4D~\cite{guo2025dist}  & \underline{32.44} & 15.63 & \underline{26.80} & \underline{60.32} & \underline{21.69} & \underline{10.99} & \underline{0.56} & 1.11 & 1.91 \\
Ours~\cite{guo2025dist}  & \textbf{34.44} & \textbf{18.06} & \textbf{28.80} & \textbf{62.32} & \textbf{23.69} & \textbf{12.99} & \textbf{0.55} & \textbf{1.09} & \textbf{1.85} \\
\bottomrule
\end{tabular}
}
\label{tab_planning}
\end{center}
\end{table}

\begin{table*}[ht]
  \centering
      \caption{Comparison of different approaches on domain generalization protocols.}
    \renewcommand\arraystretch{1}
    \setlength\tabcolsep{12pt}
    \scalebox{0.85}{\begin{tabular}{c|ccccc}
    \toprule
    Waymo $\rightarrow$ \rm{nuScenes} &\multicolumn{5}{c}{Target Domain (nuScenes)} \\
    \midrule
    Method & mAP$\uparrow$ & mATE$\downarrow$ & mASE$\downarrow$ & mAOE$\downarrow$& NDS* $\uparrow$\\
    \midrule
    Oracle &0.475 &0.577 &0.177 &0.147 &0.587\\
    \midrule
    DG-BEV &0.303 &0.689 &0.218 &0.171 &0.472\\
    PD-BEV &0.311 &0.686 &0.216 &0.170 &0.478\\
    \textbf{Ours}& \textbf{0.331} &\textbf{0.665}&  \textbf{0.26} &\textbf{0.161}&\textbf{0.498}\\
    \bottomrule
  \end{tabular}}  
  \label{tab:dg}
\end{table*}

\textbf{Ablation Experiments.} PhiGensis makes two key contributions: (1) 4D Generation Pipeline: As validated in the previous chapter through comparisons with the SOTA methods; (2) Stereo Forcing (SF). To verify the effectiveness of our method in long video generation, we first present quantitative results in Tab.~\ref{tab:fid_comparison} and further provide qualitative visualizations of video generation results with and without SF in Figure \ref{fig:sf}. Notably, Stereo Forcing enables diffusion models to better leverage geometric priors for learning consistency by predefining specific uncertainty metrics. We have tested three types of such metrics in our experiments: random noise, entropy of depth classification, and localization potential \cite{Park2022TimeWT}, with the corresponding ablation results detailed in Table \ref{tab_sf}. PhiGensis demonstrates that incorporating additional geometric uncertainty information helps mitigate the degradation tendency of diffusion models, which confirms the high effectiveness of our chosen metric—the localization potential, as it has been shown to capture the impacts of temporal fusion, multi-view fusion, and camera parameters on geometric estimation. Furthermore, our method theoretically supports multiple approaches to integrating indicators of geometric adequacy and geometric uncertainty; here, we focus on validating the feasibility of this integration framework.

\textbf{Downstream applications.} Beyond evaluating the visual quality of generated images and videos, we further assess their utility in downstream autonomous driving tasks—specifically perception and open-loop planning—using the UniAD framework \cite{hu2023planning}. Quantitative results in Table \ref{tab_planning} demonstrate that the high-fidelity videos generated by our method achieve performance that is well-aligned with original ground truth data, highlighting the potential of synthetic data to support practical downstream applications. A comprehensive analysis of Table \ref{tab_planning} reveals our method’s superiority across perception, segmentation, and planning tasks. Our method maintains the lowest L2 errors at all horizons. This outperforms baselines like DA (1.89 at 3.0m) and Dist-4D (1.91 at 3.0m), validating that synthetic videos preserve meaningful motion and scene dynamics for planning.

To verify our algorithm’s capability in 3D scene reconstruction, its ability to synthesize novel view. We focus on the task of adapting to novel vehicle models. Introducing a new vehicle model often alters camera parameters, including intrinsic parameters (e.g., camera type, focal length) and extrinsic parameters (e.g., camera placement, orientation) \cite{wang2023towards, lu2025towards}. A robust 4D reconstruction model should render images with diverse camera parameters to mitigate overfitting to specific vehicle-mounted camera configurations. 

On the Waymo dataset, we rendered images with randomly sampled intrinsic parameters and synthesized novel views with random extrinsic variations, treating these rendered outputs as augmented data. Following the protocols in \cite{wang2023towards, lu2025towards}, we trained the BEVDepth model on a combined dataset of original Waymo data and our rendered augmented data, using PD-BEV \footnote{\url{https://github.com/EnVision-Research/Generalizable-BEV}} as the baseline framework. Notably, the rendered images also underwent standard augmentation pipelines (e.g., resizing, cropping).

Table \ref{tab:dg} presents the domain generalization results when transferring from Waymo to nuScenes. Our method achieves the highest performance across all metrics on the target nuScenes domain. A critical insight is that joint augmentation of intrinsic and extrinsic parameters drives this improvement, whereas augmenting only intrinsic parameters yields limited gains. This is because virtual depth (derived from our 4D reconstruction) already addresses intrinsic parameter variations effectively, while extrinsic augmentation further enables the model to learn robust stereo relationships between cameras, which is the key to generalizing across novel vehicle.

\section{Conclusions}

This paper introduces PhiGenesis, a unified framework for geometry-aware 4D scene generation that addresses the limitations of existing simulation methods for autonomous driving. While prior approaches either rely on predefined ego trajectories or require per-scene optimization, PhiGenesis enables scalable and temporally consistent 4D scene synthesis by combining a feed-forward reconstruction stage with a trajectory-conditioned video diffusion model. A novel range-view adapter extends a frozen video VAE to reconstruct 4D Gaussian representations from monocular or multi-view inputs without scene-specific tuning. To ensure geometric fidelity during generation, the framework incorporates Stereo Forcing, an uncertainty-guided conditioning technique that corrects inconsistencies arising from depth prediction errors. Together, these innovations allow PhiGenesis to produce realistic, controllable, and structurally coherent 4D driving scenes, advancing the capabilities of simulation for autonomous driving systems. By supporting trajectory-conditioned generation, multi-view inputs and outputs, and uncertainty-aware refinement, PhiGenesis offers a scalable solution that bridges the gap between generative video models and 3D-aware scene synthesis. Experimental results demonstrate that PhiGenesis outperforms existing methods in terms of visual quality, temporal consistency, and geometric accuracy, making it a strong candidate for realistic data generation in safety-critical autonomous driving applications.

\begingroup
\let\,\relax
\bibliographystyle{IEEEtran}
\bibliography{IEEEabrv,PhiGensis}
\endgroup

\begin{IEEEbiography}[{\includegraphics[width=1in,height=1.25in,clip,keepaspectratio]{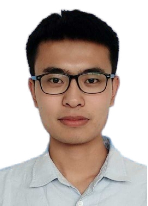}}]{Hao Lu} is a PhD candidate in Computer Science at the Hong Kong University of Science and Technology, Guangzhou, specializing in computer vision and artificial intelligence, with a particular focus on autonomous driving technology. Prior to pursuing his PhD, he obtained a master's degree from the Chinese Academy of Sciences. He has published several papers in journals and conferences such as TPAMI, TIP, CVPR, ICCV, and ECCV etc.
\end{IEEEbiography}

\begin{IEEEbiography}[{\includegraphics[width=1in,height=1.25in,clip,keepaspectratio]{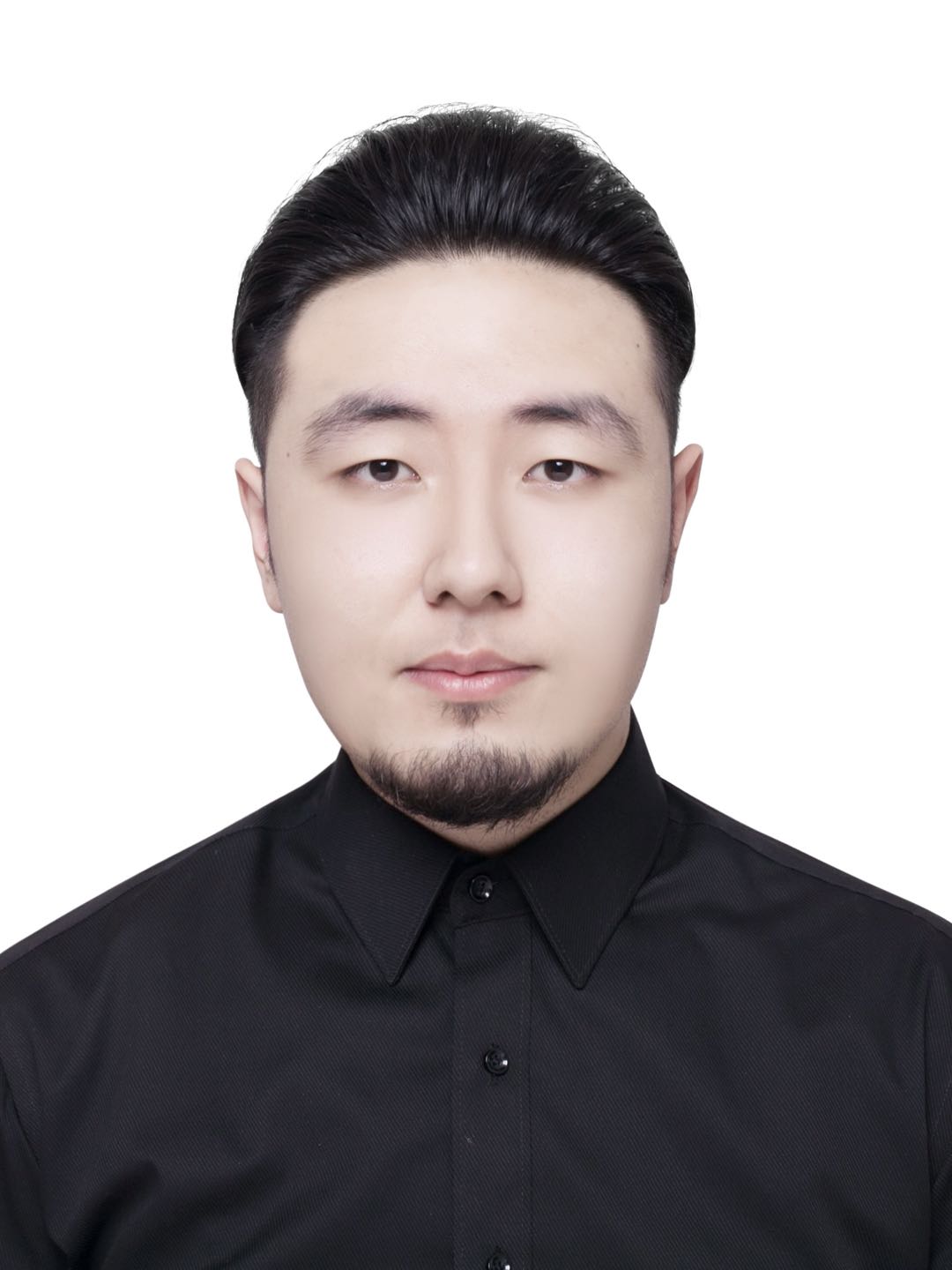}}]{Zhuang Ma} received the B.E. degree from the University of Plymouth, UK. He received the MSc degree from the University of Birmingham, UK. He is currently an engineer at PhiGent. China. His current research interests include 2D and 3D visual perception, robotics, and multi-modality content generation.
\end{IEEEbiography}

\begin{IEEEbiography}[{\includegraphics[width=1in,height=1.25in,clip,keepaspectratio]{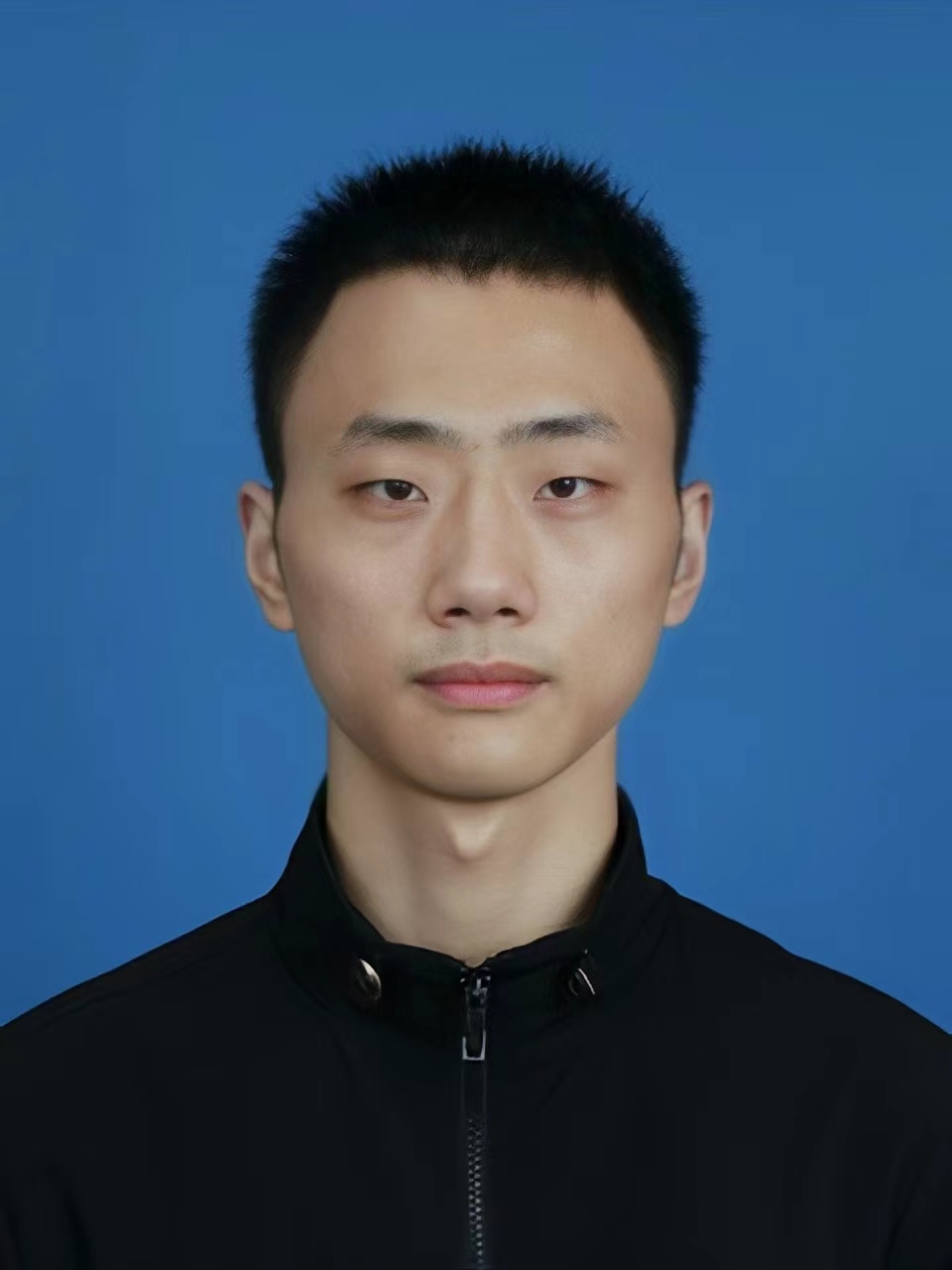}}]{Guangfeng Jiang} received the B.S. degree in communication engineering from the Shandong University, Jinan, China, in 2021. He is currently pursuing the Ph.D. degree in information and communication engineering with the University of Science and Technology of China, Hefei. His research interests include autonomous driving and 3D computer vision.
\end{IEEEbiography}

\begin{IEEEbiography}[{\includegraphics[width=1in,height=1.25in,clip,keepaspectratio]{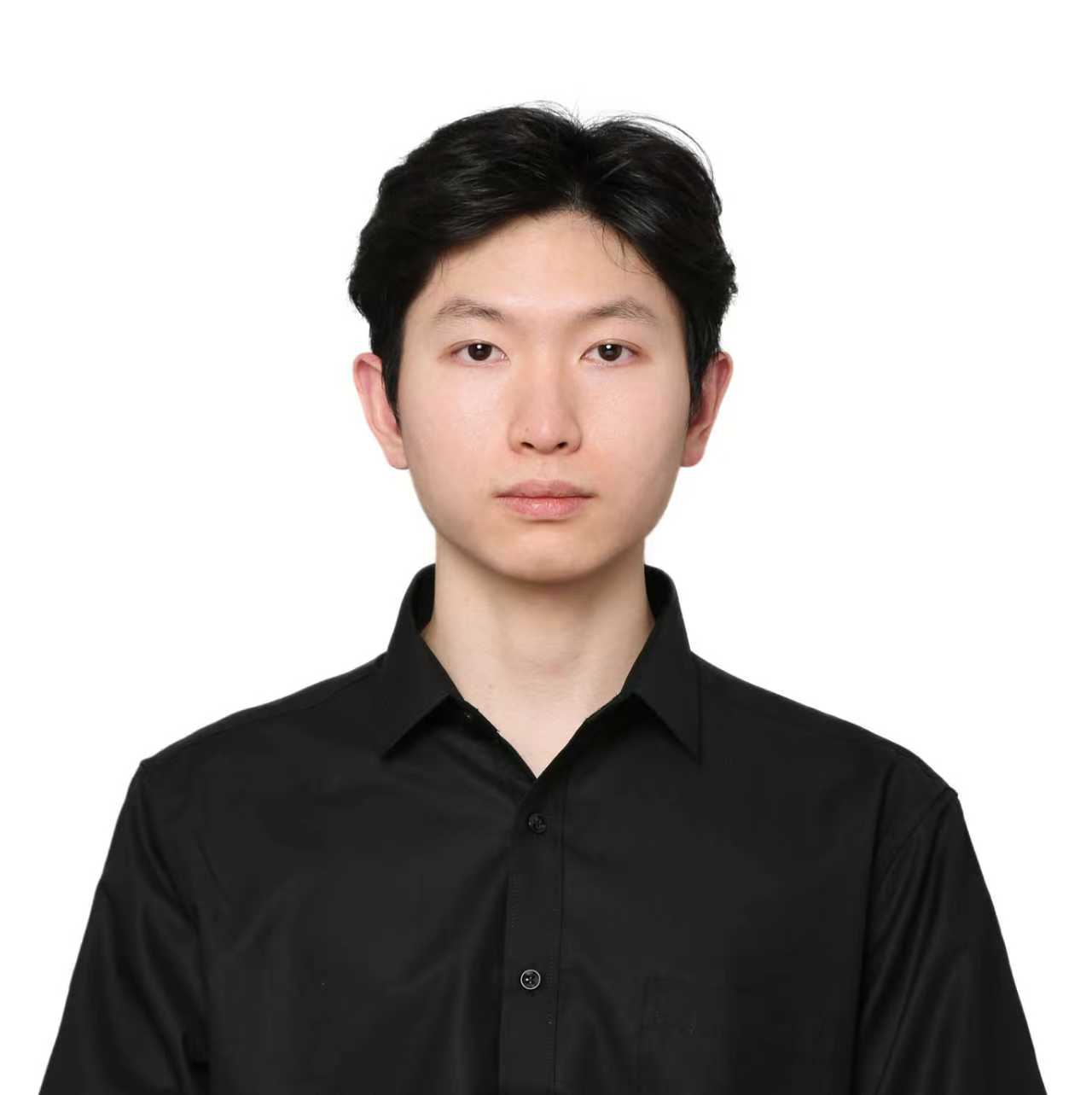}}]{Wenhang Ge} is a PhD candidate in Computer Science at the Hong Kong University of Science and Technology, Guangzhou, specializing in computer vision and artificial intelligence, with a particular focus on 3D vision and AIGC. Prior to pursuing his PhD, he obtained a master’s degree from Sun Yat-sen University. He has published several papers in journals and conferences such as TPAMI, ICCV, CVPR, and ECCV etc.
\end{IEEEbiography}

\begin{IEEEbiography}[{\includegraphics[width=1in,height=1.25in,clip,keepaspectratio]{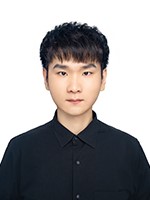}}]{Bohan Li} (Student Member, IEEE) received the B.E. degree from the School of Control Engineering, Northeastern University (NEU),
Shenyang, China, in 2019. He received the M.E. degree from the School of Control Science and Engineering, South China University of Technology
(SCUT), Guangzhou, China, in 2022.
He is currently pursuing the Ph.D. degree in Shanghai Jiao Tong University (SJTU) and Eastern Institute of Technology (EIT). His research interests include 3D visual perception, robotics, and multi-modality content generation.
\end{IEEEbiography}

\begin{IEEEbiography}[{\includegraphics[width=1in,height=1.25in,clip,keepaspectratio]{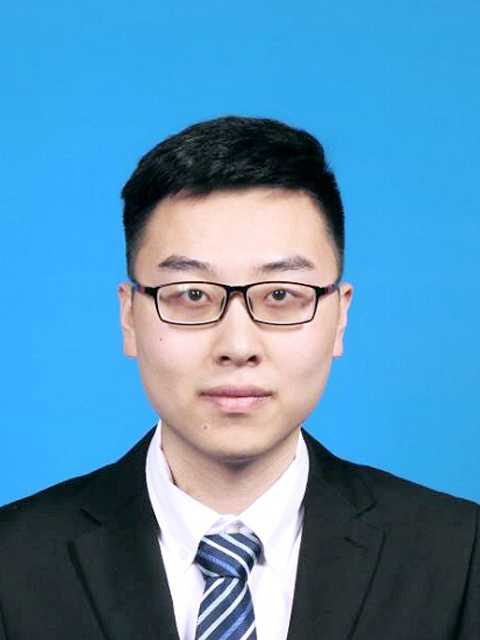}}]{Wenzhao Zheng} received the B.S. degree and the Ph.D. degree from the Department of Physics and Department of Automation, Tsinghua University, China, in 2018 and 2023, respectively. His current research interests include computer vision, deep learning, and representation learning. He has authored more than 30 papers in IEEE Transactions on Pattern Analysis and Machine Intelligence, IEEE Transactions on Image Processing, CVPR, ICCV, and ECCV. He serves as a regular reviewer member for a number of journals and conferences, e.g., IEEE Transactions on Pattern Analysis and Machine Intelligence, IEEE Transactions on Image Processing, IEEE Transactions on Biometrics, Behavior, and Identity Science, ACM Transactions on Intelligent Systems and Technology, CVPR, ICCV, ECCV, IJCAI, ICME, and ICIP.
\end{IEEEbiography}

\begin{IEEEbiography}[{\includegraphics[width=1in,height=1.25in,clip,keepaspectratio]{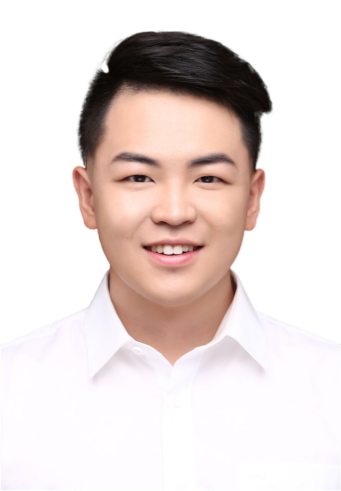}}]{Yunpeng Zhang} received the B.Sc. and M.Sc. degrees in Automation from Tsinghua University in 2019 and 2022, respectively. He is currently an algorithm engineer at PhiGent Robotics Co., LTD. His main research interests include monocular 3D object detection, multi-camera based 3D object detection, vision-based occupancy prediction, and end-to-end autonomous driving. 
\end{IEEEbiography}


\begin{IEEEbiography}[{\includegraphics[width=1in,height=1.25in,clip,keepaspectratio]{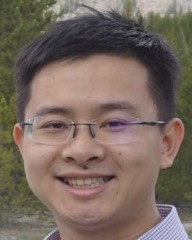}}]{Yingcong Chen} is currently an assistant professor at the Artificial Intelligence Trust, the Hong Kong University of Science and Technology (Guangzhou). He received his PhD degree from
the Chinese University of Hong Kong. He was a postdoctoral associate at the Computer Science and Artificial Intelligence Lab (CSAIL), Massachusetts Institute of Technology in 2022. He obtained the Hong Kong PhD Fellowship in 2016. He is currently an assistant professor at the Artificial Intelligence Trust, the Hong Kong University of Science and Technology (Guangzhou). He serves as a reviewer for TPAMI, IJCV, TIP, CVPR, ICCV, ECCV, BMVC, IJCAI, AAAI, etc. His research interest includes deep learning, image generation and editing, generative adversarial networks, etc.
\end{IEEEbiography}

\end{document}